\documentclass[runningheads]{llncs}
\usepackage[T1]{fontenc}
\usepackage{cite}
\usepackage{amsmath,amssymb,amsfonts}
\usepackage{algorithmic}
\usepackage{mdwtab}
\usepackage{booktabs}
\usepackage{textcomp}
\usepackage{subfigure}
\usepackage{eqparbox}
\usepackage{xcolor}
\usepackage{graphicx}
\begin{document}
\sloppy
\title{Adversarial Example Defense via Perturbation
Grading Strategy \thanks{ Zhaoxia Yin is corresponding author. Email: \email{zxyin@cee.ecnu.edu.cn}}}
\author{Shaowei Zhu\inst{1}\orcidID{0000-0003-4567-0588}\and
Wanli Lyu\inst{1}\and
Bin Li\inst{2} \and
Zhaoxia Yin{*}\inst{3}\orcidID{0000-0003-0387-4806} \and
Bin Luo\inst{1}\orcidID{0000-0001-5948-5055}
}
\authorrunning{Zhu et al.}
\institute{Anhui Provincial Key Laboratory of Multimodal Cognitive Computation Anhui University
Hefei, China\\ \email{zhusw520@gmail.com,wanly\_lv@163.com,ahu\_lb@163.com}
\and Guangdong Key Laboratory of Intelligent Information Processing and Shenzhen Key Laboratory of Media Security Shenzhen University 
Shenzhen, China\\ \email{libin@szu.edu.cn}
\and School of Communication \& Electronic Engineering East China Normal University
Shanghai, China\\ \email{zxyin@cee.ecnu.edu.cn}
}
\maketitle              % typeset the header of the contribution
\begin{abstract}
Deep Neural Networks have been widely used in many fields. However, studies have shown that DNNs are easily attacked by adversarial examples, which have tiny perturbations and greatly mislead the correct judgment of DNNs. Furthermore, even if malicious attackers cannot obtain all the underlying model parameters, they can use adversarial examples to attack various DNN-based task systems. Researchers have proposed various defense methods to protect DNNs, such as reducing the aggressiveness of adversarial examples by preprocessing or improving the robustness of the model by adding modules. However, some defense methods are only effective for small-scale examples or small perturbations but have limited defense effects for adversarial examples with large perturbations. This paper assigns different defense strategies to adversarial perturbations of different strengths by grading the perturbations on the input examples. Experimental results show that the proposed method effectively improves defense performance. In addition, the proposed method does not modify any task model, which can be used as a preprocessing module, which significantly reduces the deployment cost in practical applications.
\end{abstract}
\section{Introduction}
Deep neural networks (DNNs) have achieved widespread success in modern life, including image classification \cite{lanchantin2021general}, medical image segmentation \cite{ji2021learning}, and vehicle detection \cite{srivastava2021survey}. Research \cite{szegedy2014intriguing} has shown that attackers can add carefully crafted tiny perturbations to normal examples to mislead the model into making bad decisions. The new input generated by deliberately adding tiny perturbations to normal examples is called adversarial examples, which can lead to misjudgment of the model and cause great harm.

However, existing studies have shown that there are also adversarial examples in real physical scenarios, so there is a significant safety problem in the practical application of DNN, such as automatic driving \cite{quinonez2021shared} and face recognition \cite{dong2019efficient}. The safety requirements are higher in such case task scenarios.

Researchers have proposed various defense methods to reduce the impact of such adversarial perturbations, including adversarial training \cite{ganin2016domain,shafahi2019adversarial}, defense distillation \cite{papernot2016distillation}, and preprocessing of input transformations \cite{liu2019feature,yin2020war,jia2019comdefend}. Among them, adversarial training and defensive distillation require retraining or modification of the classifier. At the same time, input transformation-based methods focus on denoising/transforming the input before feeding it into the classifier, making it easier to deploy in practical applications. For these reasons, many input transformation-based methods have emerged in recent years. For example, ComDefend \cite{jia2019comdefend}, Deep Image Prior (DIP) \cite{ulyanov2018deep}, and DIPDefend \cite{dai2022deep} directly reconstruct adversarial examples into normal images. Similarly, DefenseGAN \cite{samangouei2018defense} uses generative adversarial networks (GANs) to remove the effects of adversarial perturbations. However, these defense methods heavily depend on the dataset size and training time for training models and are computationally expensive, limiting their real-life applications. Therefore, some researchers \cite{liu2019feature,yin2020war} turn to study how to denoise through image processing techniques.

\begin{figure*}[!ht]
    \centering
    \includegraphics[width = 4.8 in]{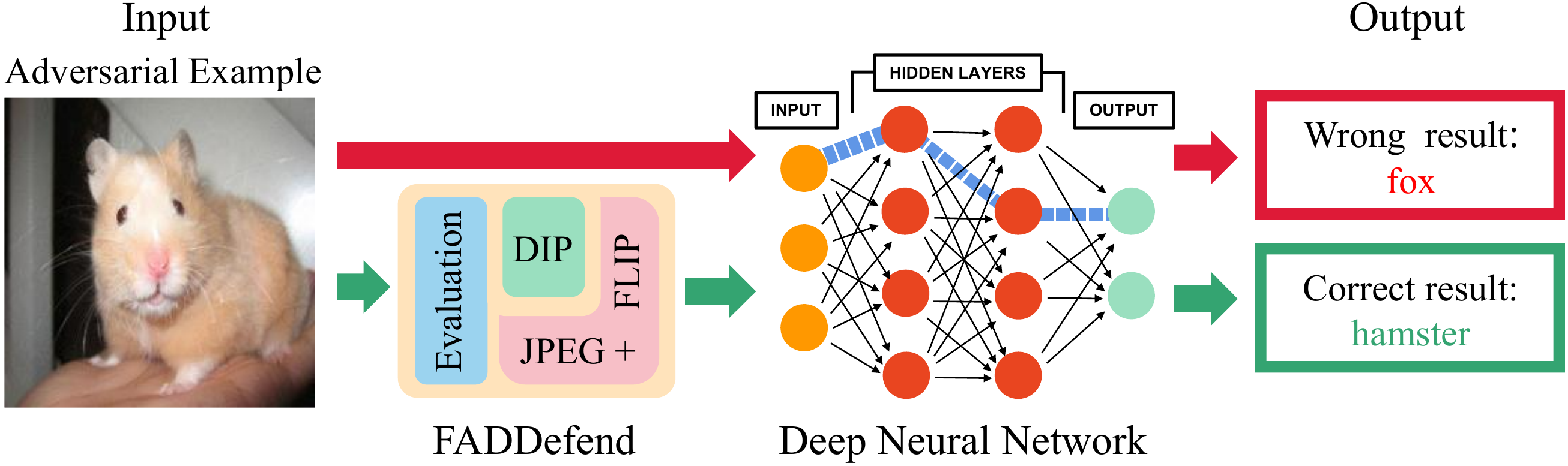}
    \centering
    \caption{\centering An example of the proposed FADefend. It removes the perturbations in the adversarial examples before feeding them into the classifier.}
    \label{fig1}
\end{figure*}

Yin et al. \cite{yin2020defense} found that the image compression method based on image processing technology can remove the perturbation from the small-perturbed image, and combined with the mirror flip, the defense effect can be improved. However, as the adversarial perturbation increases, the defense effect becomes worse. Furthermore, DIP \cite{ulyanov2018deep} can be obtained from a single low-dimensional robust feature extracted from the input image to reconstruct the image without additional training cost. The reconstructed images of this method can significantly reduce the aggressiveness of large-perturbed adversarial examples. Therefore, we propose a perturbation grading strategy-based defense method called FADDefend, which classifies adversarial perturbations and performs different operations on different levels of perturbations to improve the defense performance. The proposed method's adversarial example defense process is shown in Fig. \ref{fig1}.

The main contributions of our paper are:

\begin{itemize}
\item[$\bullet$] We propose an effective adversarial defense method that takes different defensive operations on adversarial examples with different perturbations. 

\item[$\bullet$] The proposed method uses a perturbation grading strategy to divide adversarial examples into large and small-perturbed examples.

\item[$\bullet$] Experimental results show that the proposed method not only improves the performance of the adversarial defense but also reduces the computational cost.
\end{itemize}

\section{Related Works}
\subsection{Adversarial Attack Methods}

Adversarial examples are input images to which tiny perturbations are maliciously added to fool the neural network classifier. Common adversarial attack algorithms can be divided into attack methods based on white-box settings and attack methods based on black-box settings. Under the white-box attack, the attacker can fully understand the structure and parameters of the target model, while under the black-box attack, the attacker knows nothing about the relevant information of the target model.

\subsubsection{White-box Attacks}

Szegedy et al. generate adversarial examples using L-BFGS \cite{szegedy2014intriguing}, and Goodfellow et al. proposed a method to generate adversarial examples with only a one-step attack called the Fast Gradient Sign Method (FGSM) \cite{goodfellow2014explaining}. In order to improve the attack performance of FGSM, iteration-based multi-step attack methods are proposed, such as the Basic Iterative Method (BIM) \cite{kurakin2018adversarial}, Momentum-based Iterative FGSM (MIFGSM) \cite{dong2018boosting}, and Projected Gradient Descent (PGD) \cite{madry2018towards}. DeepFool \cite{moosavi2016deepfool} attack method proposed to generate adversarial examples by finding the minimum perturbation on the hyperplane. The C\&W attack \cite{carlini2017towards} is another way to find adversarial examples through optimization. JSMA \cite{papernot2016limitations} is a sparse attack method that only modifies a small number of pixels. Athalye et al. proposed an attack method called Backward Pass Differentiable Approximation \cite{athalye2018obfuscated} by approximating the gradient of the defense to break this stochastic defense method.

\subsubsection{Black-box Attacks}

Black-box attacks usually generate black-box adversarial examples on surrogate models and then exploit the transferability of adversarial examples to attack the target model. Various methods are proposed to improve the adversarial transferability of black-box attacks, such as query-based attacks \cite{dong2019efficient,wang2022attention} and transfer-based attacks \cite{dong2019evading}.

\subsection{Adversarial Defense Methods}

Many adversarial defense methods have been proposed, including adversarial training and input transformation methods. The first method improves the robustness of the model by adding some adversarial examples to the normal training dataset, but this method slightly reduces the accuracy of normal examples. The second method modifies the input examples by preprocessing before entering the model to eliminate the adversarial perturbations in the examples. Classic digital image processing techniques, such as color depth reduction \cite{wang2021adversarial}, image stitching \cite{ding2021delving}, and JPEG compression \cite{liu2019feature}, are used to improve model accuracy. However, these methods perform unsatisfactorily in defending against large-perturbed adversarial examples.

Later, model-based image reconstruction methods are proposed, including 1) denoising adversarial examples and 2) restoring them to clean images through CNN networks. Liao et al. proposed a high-level representation-guided denoiser \cite{liao2018defense} to remove adversarial perturbation. However, it requires a large dataset and more iterations to train a denoising model that transforms adversarial examples into clean images, limiting its practical scope.

Our work focuses on improving the unsatisfactory robustness of these defenses to large-perturbed adversarial examples. Since previous defense methods have shown better defense against small-perturbated adversarial examples, this problem can be solved by transforming large-perturbed images into small-perturbed images through the DIP \cite{ulyanov2018deep} network.

\section{Proposed Method}

\begin{figure*}[!ht]
    \centering
    \includegraphics[width = 4.7 in]{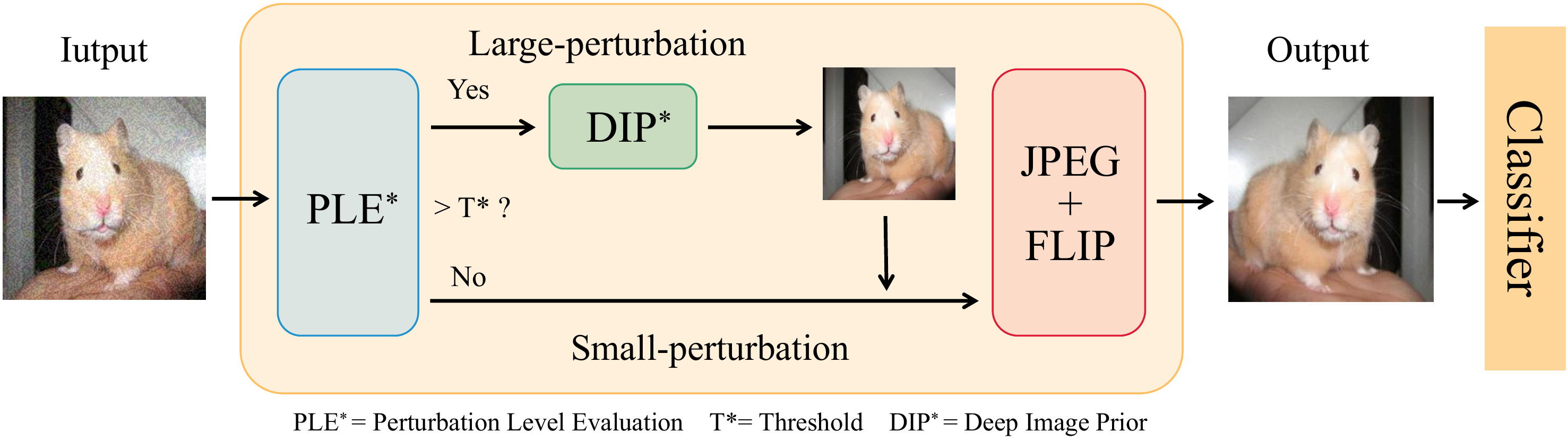}
    \centering
    \caption{\centering FADDefend defense framework.}
    \label{fig2}
\end{figure*}

This paper proposes a new defense method FADDefend. FADDefend is divided into three modules. The first module is an evaluation module used to evaluate the noise level in the example. The second module is a JPEG compression and mirror flip module used to process the small-perturbed examples after perturbation grading. The third module is the DIP reconstruction module used to process large-perturbed examples after perturbation grading. Specifically, when
an example is fed into FADDefend, the evaluation module evaluates its perturbation level. If the perturbation level is less than a pre-set threshold, it is defined as a small perturbation. It is denoised using a JPEG compression algorithm with a quality factor (QF) of 95, then mirror flipped and fed into the classifier. Otherwise, it is defined as a large perturbation and sent to the DIP image restoration module, which is sent to the JPEG compression and mirror flip module after processing. Fig. \ref{fig2} shows the framework of FADDefend.

The basic idea of the proposed method to resist adversarial perturbation is introduced in this section. Then the pros and cons of the JPEG compression and DIP defense methods are revealed. Finally, the proposed method combines the advantages of the two defense methods through the perturbation level evaluation module to achieve a better defense effect.

\subsection{Image perturbation level evaluation}

Defenders assume a known perturbation level and conduct targeted defenses, but this assumption is unrealistic. When the model is deployed, it is attacked by an unknown perturbation strength, and its perturbation level cannot be judged based on the input image. Therefore, this problem can be solved by evaluating the perturbation level of the input image by blind image perturbation level evaluation.

Liu et al. proposed a blind image perturbation level assessment algorithm \cite{liu2013single} to select low-level patches without high-frequency information from noisy images. The Principal Component Analysis technique estimates the perturbation level based on the selected patches. The eigenvalues of the image gradient covariance matrix are used as the standard to measure the texture intensity. A stable threshold is selected to distinguish the perturbation level through an iterative method. 

\begin{figure}[!ht]
    \centering
    \includegraphics[width = 3.4 in]{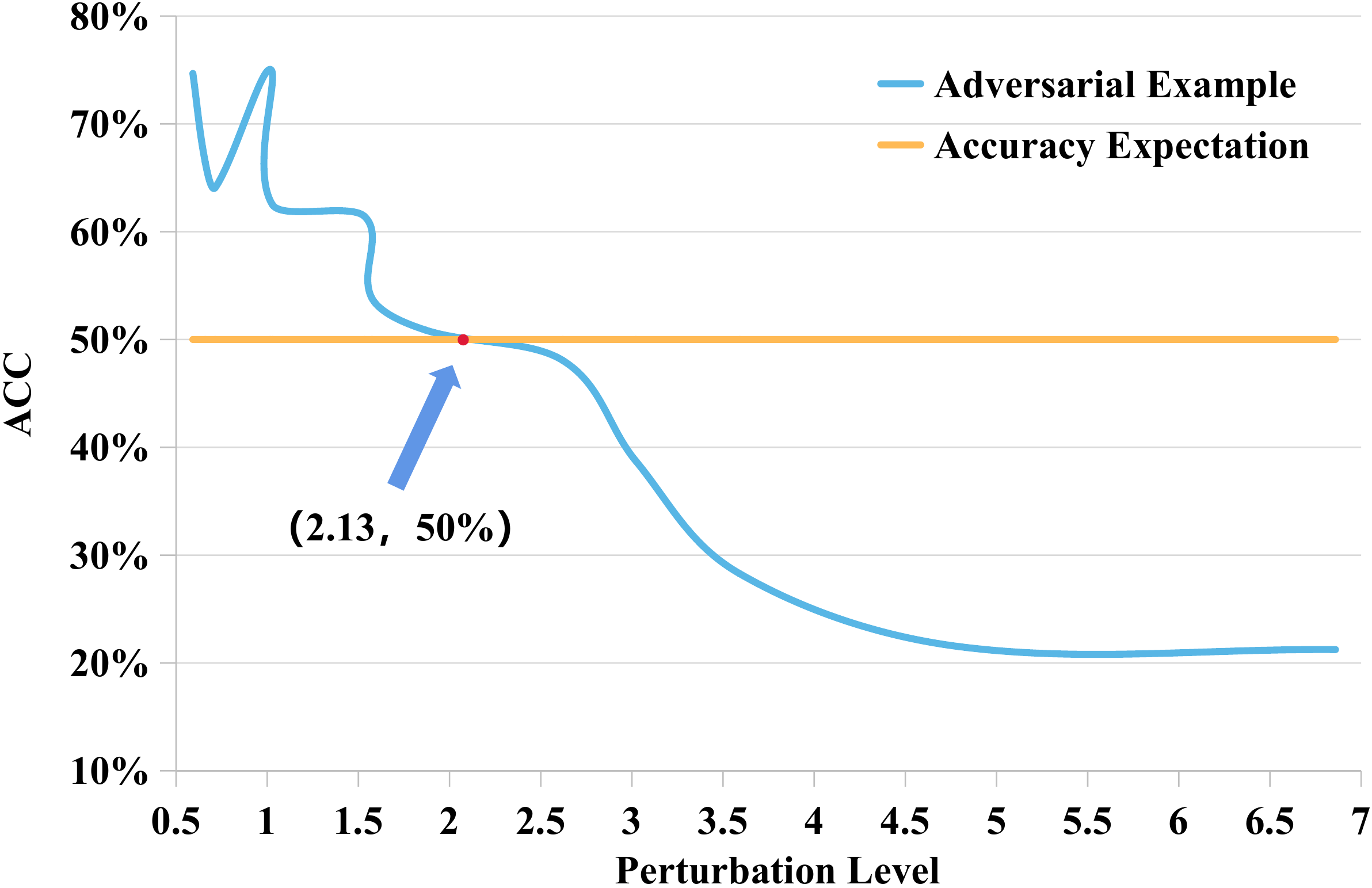}
    \centering
    \caption{Different thresholds can be chosen by the intersection of expected accuracy and adversarial example defense accuracy.}
    \label{fig3}
\end{figure}

In this section, the experimental examples are composed of 500 examples from the data set and the expected accuracy of the default is set
to 50\% : A choice that balances detection accuracy and defense time consumption. Therefore, the threshold is chosen by comparing the defense accuracies of adversarial examples with different perturbation levels, as shown in Fig.\ref{fig3} (Threshold is 2.13). Those smaller than the threshold are defined as small-perturbed adversarial examples, and those greater than the threshold are defined as large-perturbed adversarial examples.

The perturbation strength can be divided into different intervals by selecting different thresholds, convenient for subsequent selection of different image processing modules.

\subsection{Defense methods based on image processing technology}
\begin{figure*}[!ht]
    \centering
    \includegraphics[width = 4.5 in]{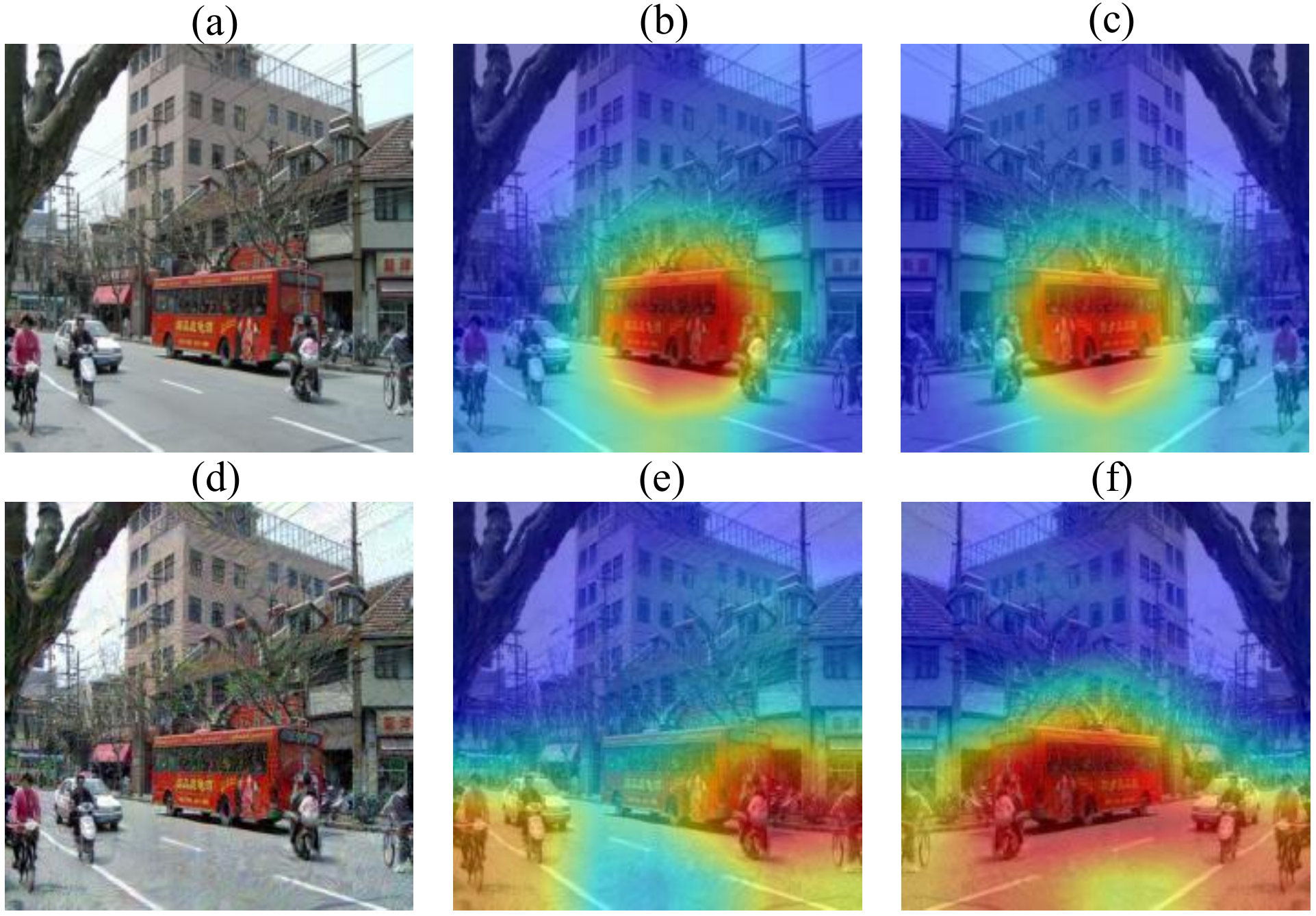}
    \centering
    \caption{(a) original example; (d) corresponding adversarial example. (b)(c)(e)(f) used class activation mapping of the images. (b) original example-bus; (c) fliped original example-bus; (e) adversarial example-bike; (f) fliped adversarial example-bus. The redder the class activation mapping of the image, the more the model pays attention to this area.}
    \label{fig4}
\end{figure*}

\begin{figure*}[!t]
    \centering
    \subfigure[]{
    \centering
        \begin{minipage}[]{0.4\textwidth}
            \includegraphics[width=1\textwidth]{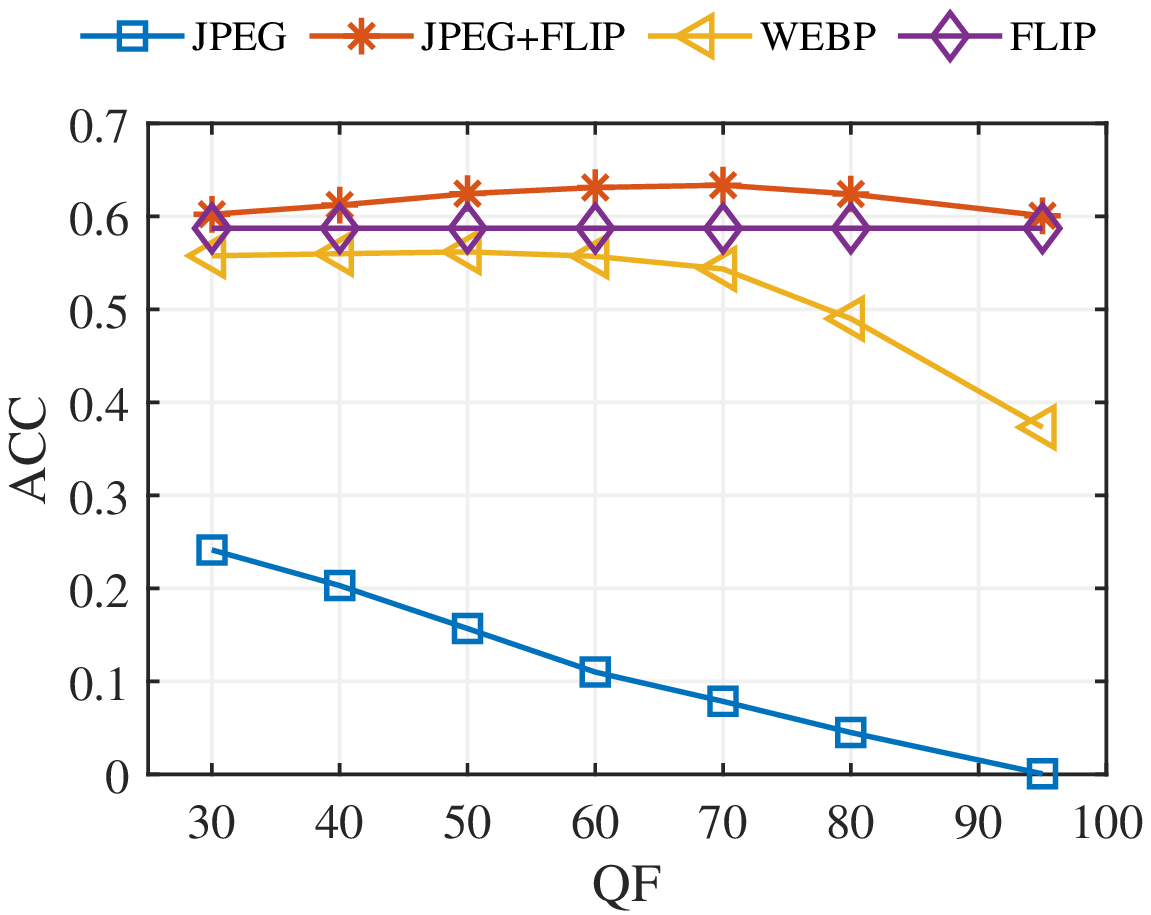} 
        \label{a}
        \end{minipage}
    }
    \subfigure[]{
    \centering
        \begin{minipage}[]{0.4\textwidth}
            \includegraphics[width=1\textwidth]{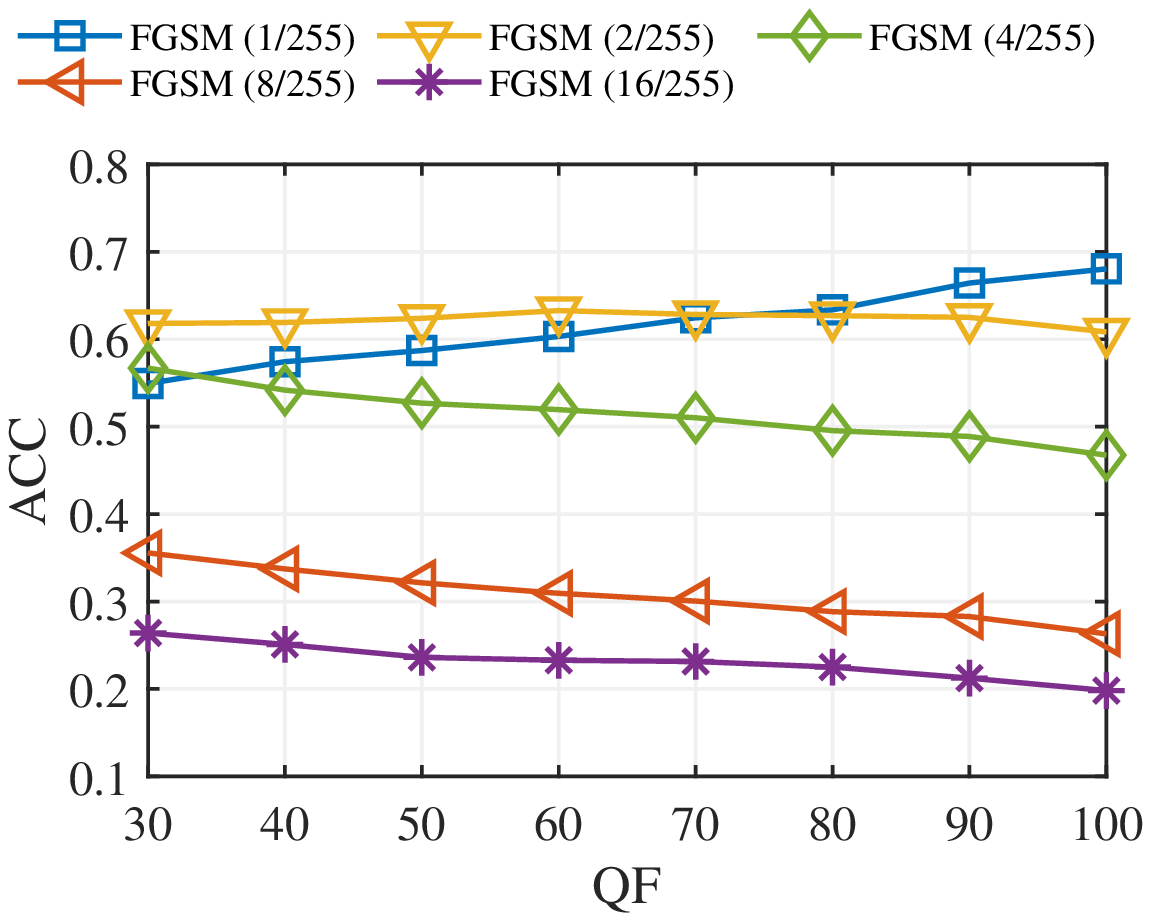} 
            \label{b}
        \end{minipage}
        }
    \caption{ Comparison of defense effects of preprocessing methods under different QFs. (a) Defense accuracy of various preprocessing methods under FGSM (2/255), (b) Defense accuracy using JPEG compression combined with mirror flip under different perturbations.}
    \label{fig5}
\end{figure*}

Most existing defense methods cannot defend well against large-perturbed adversarial examples. For example, the standard JPEG compression algorithm can remove the high-frequency information in the image well to retain the low-frequency information. Just as research \cite{liao2018defense} shows that adversarial perturbations can be viewed as high-frequency information with a specific structure so that this image compression can achieve a specific denoising effect. However, as the perturbation strength increases, the adversarial perturbation may enter the low-frequency range. Even after removing some high-frequency information, the adversarial perturbation still exists. 

We strengthen JPEG's defense against small perturbations by breaking the adversarial structure by mirror flipping \cite{yin2020defense}. As shown in Fig. \ref{fig4}, the model's attention is observed by flipping the original and adversarial examples and using the class activation map. For the original example, the model's attention is focused on the object's main part before and after the flip. At the same time, for the adversarial example, the model is distracted, and the flipped adversarial example can correct the model's attention so that it can be classified correctly.

Fig. \ref{a} shows the defense accuracy of different defense methods under the QFs. As the QF increases, the compression rate decreases, reducing the defense effect of JPEG compression against adversarial examples. In contrast, JPEG compression combined with mirror flip has better defense accuracy and stability. Fig. \ref{b} shows that the defense of JPEG compression combined with mirror flip is unsatisfactory under the large-perturbation (FGSM, $\epsilon$= 8/16).

\begin{figure*}[!ht]
    \centering
    \includegraphics[width = 4.8 in]{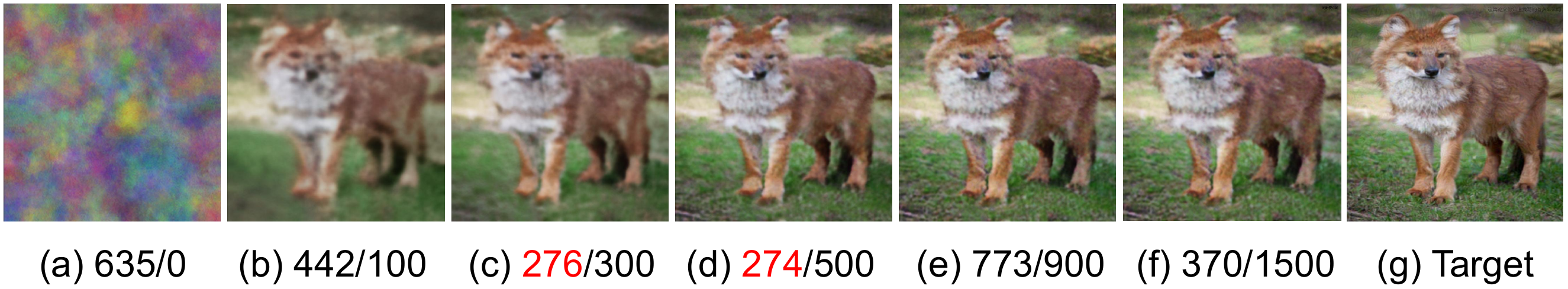}
    \centering
    \caption{ The above figures show the image reconstruction process of DIP and the image classification results (predicted category ID/iteration number) at different stages. The Top-5 predictions containing the correct label are marked in red (true labels: 274). PGD (8/255) is used to generate untargeted adversarial examples.}
    \label{fig6}
\end{figure*}

The compressed image loses semantic information by using a smaller QF (the smaller the QF, the worse the image quality), making it more difficult for the model to distinguish accurately. Although such methods defend well on small-perturbed adversarial examples and low computational cost, they cannot effectively deal with adversarial examples with large perturbations. Consider using image reconstruction methods to process large perturbations in adversarial perturbations into small perturbations, followed by JPEG and flip processing, which is introduced in the next section.

\subsection{Defense methods based on deep image priors}

Many model-based defense methods can achieve good defense results, such as image reconstructions \cite{dai2022deep}, image denoising \cite{liao2018defense}, image super-resolution \cite{ulyanov2018deep}, or GANs \cite{samangouei2018defense} recovering images for large-perturbed adversarial examples.

However, these methods require image reconstruction by learning a large amount of prior knowledge of external training data during the training stage. These defense methods consume a lot of computational and hardware costs. The CNN model itself has specific image prior capability. A clean image can also be reconstructed using prior internal knowledge of the image itself. Therefore, an untrained convolutional neural network is used in this module to generate denoised images through the DIP-based \cite{ulyanov2018deep} generator network. Specifically, an adversarial example $x_{adv}$ adds random noise $z$ as the input example of the generator, which is solved by the following constraints by
\begin{equation}
\min _{\theta}\left\|f_{\theta}(z)-x_{a d v}\right\|_{2}^{2},
\end{equation}
where $f$ is a randomly initialized network, random but fixed noise $z$ and adversarial examples $x_{adv}$ as learning targets, each iteration input is a fixed noise $z$,
and the parameter $\theta$ is updated by gradient descent. As the number of iterations
increases, the output is closer to $x_{adv}$.

Consistent with the conclusions drawn \cite{ulyanov2018deep}, the DIP-based image reconstruction method restores the low-frequency information of the image in the early iterative stage and restores the high-frequency information (including adversarial perturbation) as the number of iterations increases. Fig. \ref{fig6} shows the image reconstruction process of DIP and the image classification results. In the early stage, the network recovered the image's low-frequency information (main contours). The high-frequency information of the image will also be recovered in later iterations. It can be seen that the visual quality of the image is very high at this time. Unfortunately, adversarial perturbations are also recovered, leading to overfitting. Therefore, reconstructed images with few adversarial perturbations are obtained by stopping in an early stage. Finally, the DIP module transforms the large-perturbed example into a small-perturbed example. The small perturbation adversarial examples can increase the classification model's defense accuracy through JPEG and flip processing.

\section{Experiments}
\subsection{Experimental Settings}

Our experiments are tested on the ImageNet dataset. 10,000 test images are selected randomly from the whole test set, and 8,850 clean example sets are screened for experiments. ResNet152 \cite{he2016identity} is used to test robustness against various attack strengths of FGSM \cite{goodfellow2014explaining}, PGD \cite{madry2018towards}, BIM \cite{kurakin2018adversarial}, and MIFGSM [19] ($\epsilon$ = 2/4/8/16, using the $L_{\infty}$ distance metric). ResNet152 \cite{he2016identity} and VGG19 \cite{simonyan2015very} are used to test cross-model defense robustness.

The DIP module adopts the U-Net \cite{ronneberger2015u} structure as the generative network and skip-connections to connect outputs of layers with the same spatial dimension. Specifically, LeakyReLU is used as the nonlinear activation function. Moreover, convolution operation and bilinear interpolation are used for downsampling and upsampling. An adversarial example is input as the target image. The generator network parameters are randomly initialized, random noise is used as input, and the target image is reconstructed by gradient descent training, stopping after 400 iterations.

\subsection{Comparison with Image Compression Defenses}

FADDefend combines image compression, mirror flip, and reconstruction modules with perturbation level evaluation modules. Since the image reconstruction module has a better processing effect on the large-perturbed adversarial examples, the image compression method is used as an auxiliary. It makes up for the disadvantage of the image compression defense method that the defense effect is unsatisfying when dealing with large-perturbed adversarial examples.
\begin{figure}[!ht]
    \centering
    \includegraphics[width = 4.3 in]{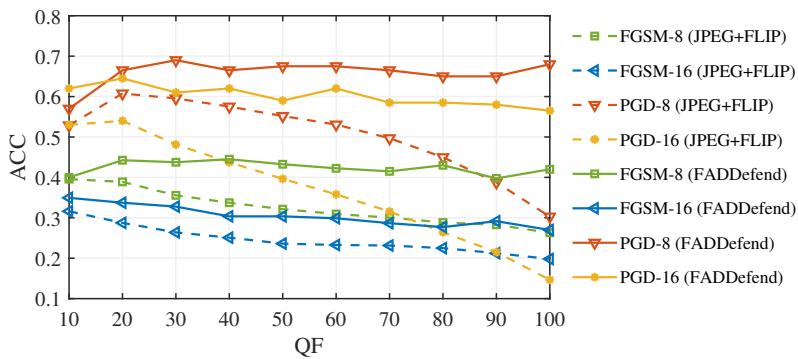}
    \centering
    \caption{ The dotted part is the defense method of JPEG Compression combined with mirror flip. The solid part is the FADDefend defense method, showing the recognition accuracy curves under different QFs.}
    \label{fig7}
\end{figure}

As shown in Fig. \ref{fig7}, compared with the defense method of JPEG image compression combined with mirror flip, FADDefend has better performance under different QFs and maintains stable defense performance with the increase of QF.

\begin{table*}[!ht]
    \centering
    \caption{Accuracy results (\%) of defense methods on ResNet152 under attacks of different types and strengths (ImageNet, $\epsilon$ = 2/4/8/16)}
    \label{table1}
    \begin{tabular}{m{2.3cm}<{\raggedright}m{1cm}<{\centering}m{2cm}<{\centering}m{2cm}<{\centering}m{2cm}<{\centering}m{2cm}<{\centering}}
    \toprule
    \textbf{Method} & \textbf{Clean} & \textbf{FGSM}  & \textbf{MIFGSM} & \textbf{PGD} & \textbf{BIM}\\
    \midrule
    Normal  & 100 & 0 / 0 / 0 / 0 & 0 / 0 / 0 / 0 & 0 / 0 / 0 / 0 & 0 / 0 / 0 / 0 \\
    JPEG\cite{liu2019feature}& 86 & 47/45/24/13 & 74/64/40/14 & 59/64/50/37 & 56/68/56/63 \\
    War\cite{yin2020war}& 86 &60/\textbf{59}/36/21 & 79/72/53/27 & 65/68/59/51 & 61/75/63/72 \\
    ComDefend\cite{jia2019comdefend}& 94 & 32/19/ 5 / 5 & 56/30/ 9 / 1  & 44/32/ 7 / 2 & 37/48/21/23 \\
    DIPDefend\cite{dai2022deep}& 79 & 55/45/26/14 & 80/67/43/25 & 60/69/60/38 & 61/71/59/77 \\
    \textbf{FADDefend}  & \textbf{94} & \textbf{62}/55/\textbf{43}/\textbf{32} & \textbf{81}/\textbf{75}/\textbf{60}/\textbf{44} & \textbf{66}/\textbf{72}/\textbf{66}/\textbf{65} & \textbf{68}/\textbf{78}/\textbf{70}/\textbf{84} \\
    \bottomrule
    \end{tabular}
\end{table*}

\begin{table*}[!ht]
    \centering
    \caption{Accuracy results (\%) of defense methods on attacks on the VGG19, the source model is ResNet152. (ImageNet, $\epsilon$ = 4/8/16)}
    \label{table2}
    \begin{tabular}{m{2.6cm}<{\raggedright}m{1cm}<{\centering}m{2cm}<{\centering}m{2cm}<{\centering}m{2cm}<{\centering}m{2cm}<{\centering}}
    \toprule
    \textbf{Method} & \textbf{Clean} & \textbf{FGSM}  & \textbf{MIFGSM} & \textbf{PGD} & \textbf{BIM}\\
    \midrule
    Normal  & 89 & 48/40/22 & 
    66/40/23 & 
    62/42/28 & 
    70/51/60\\
    War \cite{yin2020war} & 74 & \textbf{57}/42/26& 65/54/39 & 
    52/50/37 & 
    67/56/62 \\
    ComDefend \cite{jia2019comdefend} & 86 & 
    52/38/22 & 
    70/50/31 &  
    66/47/41 & \textbf{75}/60/62 \\
    DIPDefend \cite{dai2022deep} & 86 & 
    50/44/31 & 
    69/54/46 & 
    66/\textbf{65}/44& 73/61/72 \\
    \textbf{FADDefend}  & 86 & 50/\textbf{46}/\textbf{31} & \textbf{70}/\textbf{57}/\textbf{46} & \textbf{66}/61/\textbf{44} & 73/\textbf{62}/\textbf{72} \\
    \bottomrule
    \end{tabular}
\end{table*}

\subsection{Results on Attacks of Different Types and Strengths}

To verify the effectiveness of the proposed FADDefend method, other state-of-the-art defense methods are compared, including JPEG compression \cite{liu2019feature}, War (WebP compression and resize) \cite{yin2020war}, ComDefend \cite{jia2019comdefend}, and DIPDefend \cite{dai2022deep}. The classification accuracy on ImageNet is shown in Table \ref{table1}. In contrast, the image reconstructed methods have high defense accuracies, such as DIPDefend \cite{dai2022deep} and the proposed method. Under the FGSM attack($\epsilon$= 8/16), compared with the other four methods, the proposed method improves accuracy by at least 10\%. Compared with the image compression method of ComDefend \cite{jia2019comdefend}, the defense accuracy can even exceed 20\%. The image compression methods are generally ineffective due to the large perturbation, which destroys the image structure and causes the loss of semantic information.

\subsection{Results on Migration Attacks under Different Models}

Under cross-model attacks, assume that the attacker does not have access to the specific parameters of the target classifier or defense model. Attackers can only attack by exploiting the transferability of adversarial examples. The attacker acts as a surrogate model by training a model with a different structure than the target classifier. Then, adversarial examples generated by attacking the surrogate model may also lead to misclassification of the target classifier. Adversarial examples are generated under various attack algorithms ($\epsilon$ = 4/8/16) using the ResNet152 model as surrogate and VGG19 as the target model. The test results are shown in Table \ref{table2}. It can be seen that the recognition accuracy of the VGG19 model drops significantly on the adversarial examples generated by ResNet152. The larger the perturbation, the lower the recognition accuracy, which is the same result tested on the surrogate model and verified the attack transferability of adversarial examples. The recognition accuracy of the target model under different strengths and attack algorithms is significantly improved after adopting the proposed defense method. Compared with War \cite{yin2020war}, Comdefend \cite{jia2019comdefend}, and DIPDefend \cite{dai2022deep} defense methods, the proposed method achieves the same or even better performance under various attacks. 

The data in Table \ref{table2} shows that the defense performance of the proposed method is almost equal to that of DIPDefend under specific attack strengths, but this does not mean that the proposed method has no advantages. At the same time, we also conduct comparative experiments on the running speed, and memory consumption with the image reconstruction method \cite{ulyanov2018deep} and the method \cite{dai2022deep} with the defense accuracy close to the proposed method. Randomly select 1000 images from the same dataset in Table \ref{table2} to form an example dataset for experimentation. Table \ref{table3} shows that the proposed method is less than the other two methods in running time and memory consumption because the other two methods reconstruct all the examples. Due to the number of iterations required to guarantee the generation of normal examples, the DIP method takes about 900 iterations, and DIPDefend takes about 500 iterations. We allow the generated examples to contain fewer perturbations, so only 400 iterations are needed for large-perturbed examples, and image compression algorithms are used for small-perturbed examples. Therefore, compared with other methods, the proposed method has certain advantages in defense accuracy and computational cost.

\begin{table}[!ht]
    \centering
    \caption{Compared with the running time and memory consumption of other existing image reconstruction methods.}
    \label{table3}
    \begin{tabular}{m{3cm}<{\raggedright}m{3cm}<{\centering}m{3cm}<{\centering}}
    \toprule
    Method  & Run-time (s) & Memory (MB) \\
    \midrule
    DIP \cite{ulyanov2018deep}  & 52.896 & 153.9  \\
    DIPDefend \cite{dai2022deep}  & 31.038 & 146.8  \\
    \textbf{FADDefend}  & \textbf{13.616} & \textbf{121.1} \\
    \bottomrule
    \end{tabular}
\end{table}

\section{Conclusion and Future Work}

This paper proposes an effective adversarial defense method. Different defenses are performed according to the perturbation grading strategy by evaluating the adversarial perturbation strength. The proposed method achieves better defensive performance against the ImageNet dataset under different perturbation strengths and attack algorithms, as well as in cross-model attacks. Because this classification strategy ensures defense accuracy, it also improves the running speed and reduces the computational cost.

Since the proposed method uses mirror flip, it is well defensive against real-world adversarial examples. However, in special scenarios such as numbers, mirror flip leads to semantic information changes, which need improvement.
\bibliographystyle{splncs04}
\bibliography{mybib}
\end{document}